%% file: main.tex
\documentclass[10pt,twocolumn,letterpaper]{article}

\usepackage{cvpr}              % To produce the CAMERA-READY version
\usepackage{dsfont}
\usepackage{multirow}
\usepackage{fontawesome5}
\usepackage{booktabs}
\usepackage{threeparttable}
\usepackage{amsmath}
\usepackage{amssymb}
\usepackage{amsthm}
\usepackage[table]{xcolor}

\input{preamble}
\definecolor{cvprblue}{rgb}{0.21,0.49,0.74}
\usepackage[pagebackref,breaklinks,colorlinks,allcolors=cvprblue]{hyperref}

\def\paperID{25663} % *** Enter the Paper ID here
\def\confName{CVPR}
\def\confYear{2026}

\title{GeoDiT: A Diffusion-based Vision-Language Model for Geospatial Understanding}

\author{Jiaqi Liu\textsuperscript{\rm 1,\rm 2}, Ronghao Fu\textsuperscript{\rm 1,\rm 2}\thanks{Corresponding authors.} , Haoran Liu\textsuperscript{\rm 1,\rm 2}, Lang Sun\textsuperscript{\rm 1,\rm 2}, Qipeng Wang\textsuperscript{\rm 1,\rm 2}, Bo Yang\textsuperscript{\rm 1,\rm 2}\footnotemark[1]\\
\textsuperscript{\rm 1}College of Computer Science and Technology, Jilin University, Changchun 130012, China\\
\textsuperscript{\rm 2}Key Laboratory of Symbolic Computation and Knowledge Engineering of Ministry of Education\\
{\tt\small \{liujq21,lhr24,sunlang24,wangqp24\}@mails.jlu.edu.cn, \{furh,ybo\}@jlu.edu.cn}
}

\begin{document}
\maketitle
\input{sec/0_abstract}    
\input{sec/1_intro}
\input{sec/2_relatedwork}
\input{sec/3_methodology}
\input{sec/4_experiments}
\input{sec/5_conclusion}
\input{sec/6_acknowledgments}
{
    \small
    \bibliographystyle{ieeenat_fullname}
    \bibliography{main}
}
%\input{sec/7_suppl}

% WARNING: do not forget to delete the supplementary pages from your submission 
% \input{sec/X_suppl}

\end{document}

\maketitle
\input{sec/7_suppl}
{
    \small
    \bibliographystyle{ieeenat_fullname}
    \bibliography{main}
}

% WARNING: do not forget to delete the supplementary pages from your submission 
% \input{sec/X_suppl}

%% file: sec/0_abstract.tex
\begin{abstract}
Autoregressive models are structurally misaligned with the inherently parallel nature of geospatial understanding, forcing a rigid sequential narrative onto scenes and fundamentally hindering the generation of structured and coherent outputs. We challenge this paradigm by reframing geospatial generation as a parallel refinement process, enabling a holistic, coarse-to-fine synthesis that resolves all semantic elements simultaneously. To operationalize this, we introduce GeoDiT, the first diffusion-based vision-language model tailored for the geospatial domain. Extensive experiments demonstrate that GeoDiT establishes a new state-of-the-art on benchmarks requiring structured, object-centric outputs. It achieves significant gains in image captioning, visual grounding, and multi-object detection, precisely the tasks where autoregressive models falter. Our work validates that aligning the generative process with the data's intrinsic structure is key to unlocking superior performance in complex geospatial analysis. Resources can be found in \href{https://github.com/ViTBerger/GeoDiT}{https://github.com/ViTBerger/GeoDiT}.
\end{abstract}

%% file: sec/1_intro.tex
\section{Introduction}
\label{sec:intro}

Automating the interpretation of the vast and ever-growing stream of geosptial data constitutes a pivotal challenge in the remote sensing domain. From monitoring deforestation for climate security \cite{mo2023integrated,yang2013role} to tracking infrastructure development for urban dynamics \cite{gao2020mapping,chakraborty2024large}, the ability to rapidly transform raw satellite imagery into actionable insights is critical. In response, adapting large-scale vision-language models (VLMs) on Earth observation data has emerged as the dominant and most promising paradigm. 

\begin{figure}[!t]
    \centering
    \includegraphics[width=1\linewidth]{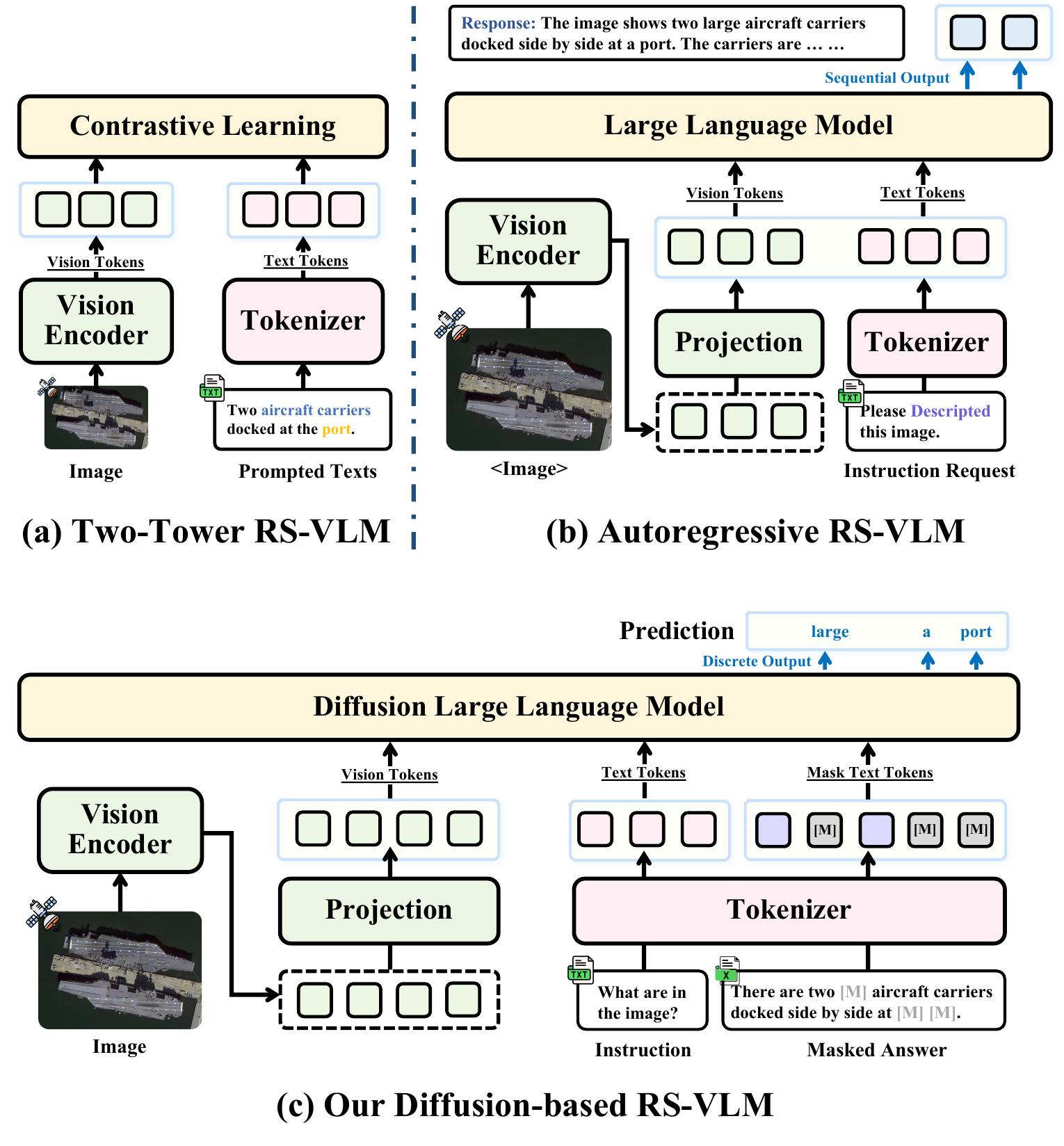}
    \caption{Conceptual comparison of text generation paradigms for geospatial understanding. (a) A two-tower model based on contrastive learning. (b) An autoregressive model generating text sequentially. (c) Our proposed diffusion-based model generating text in parallel via iterative mask prediction.}
    \label{fig:paradigm_comparison}
    \vspace{-0.7cm}
\end{figure}

This paradigm has largely evolved along two architectural lines. Initial efforts centered on aligning visual and textual representations within a shared embedding space through two-tower contrastive learning (Figure~\ref{fig:paradigm_comparison}(a)). While effective for retrieval-style tasks, these models \cite{liu2024remoteclip,zhang2024rs5m,li2023rs} lack the capacity for nuanced, generative interpretation. More recently, the frontier has shifted to autoregressive VLMs, which fuse visual features directly into a Large Language Model backbone (Figure~\ref{fig:paradigm_comparison}(b)). This architectural shift has rapidly consolidated into the prevailing methodology for generative geospatial intelligence, exemplified by a series of recent works such as GeoChat~\cite{kuckreja2024geochat}, VHM~\cite{pang2025vhm}, and EarthDial~\cite{soni2025earthdial}. 

Undeniably, the success of this autoregressive paradigm has unlocked unprecedented capabilities in geospatial intelligence on a series of tasks such as scene classification (SC), visual question answering (VQA), visual grounding (VG). However, their celebrated success in these areas, which typically demand a single, focused output, masks a crucial architectural limitation. Specifically, their strictly sequential nature is fundamentally incompatible with a coarse-to-fine synthesis, a process that first establishes a scene's global composition (such as its principal subjects for a description or the number and placement of objects for detection) before committing to the fine-grained generation of specific words or coordinates. This linear, token-by-token commitment stands in stark contrast to human cognition, where a global gist of the scene precedes the articulation of details. 

This core incompatibility, therefore, materializes as observable, systematic failures in tasks requiring structured outputs. In comprehensive scene description, the model’s generative focus becomes prematurely anchored to the first salient entity, causing it to expend its descriptive budget on initial observations while struggling to cohesively integrate other, spatially-distant concepts into a balanced narrative. In multi-object detection, the same path-dependency creates a contextual feedback loop: the generation of one bounding box pathologically influences the prediction of the next, leading to the redundant detection of a single object, in place of a systematic canvass for other distinct entities. Ultimately, these failures reveal a shared root cause: the inherent inability of a sequential process to formulate a globally consistent understanding before committing to local, token-by-token generation.

Overcoming this limitation requires a new generative paradigm with an inherently global and parallel structure. The principles of denoising diffusion models \cite{you2025lladavlargelanguagediffusion,li2025lavidalargediffusionlanguage,yang2025mmadamultimodallargediffusion,khanna2025mercury}, which have recently catalyzed breakthroughs in general-purpose visual and language generation, offer a compelling and structurally-aligned solution to this challenge. Unlike autoregressive models, which are constrained to a rigid, unidirectional generation path, diffusion models frame generation as a parallel, iterative refinement of a complete output structure. As illustrated in Figure~\ref{fig:paradigm_comparison}(c), they begin with a noisy, unstructured canvas and progressively denoise it, allowing all semantic units, be they words or coordinates, to be resolved simultaneously and co-dependently, thus building coherence at a global level from the outset.

\begin{figure*}[!t]
    \centering
    \includegraphics[width=1\linewidth]{fig/framework.pdf}
    \caption{Overview of the GeoDiT framework, illustrating the \textbf{(a)} training and \textbf{(b)} inference procedures. \textbf{(a)} During training, GeoDiT is optimized with a mask-and-predict objective. The model learns to reconstruct the original text from a randomly masked version, conditioned on a prompt and the visual features from a SigLIP-2 encoder. This follows a two-stage strategy: initial vision-language alignment by training only the MLP projector (Stage \uppercase\expandafter{\romannumeral1}), followed by end-to-end instruction tuning of the entire model (Stage \uppercase\expandafter{\romannumeral2}). \textbf{(b)} Inference is a non-autoregressive, iterative refinement process. Starting from a fully masked template, the model repeatedly predicts the full sequence and then applies a low-confidence remasking strategy. High-confidence tokens are preserved while uncertain ones are re-masked for the subsequent refinement step, a process that continues for a fixed number of iterations to produce the final output.}
    \vspace{-0.2cm}
\label{fig:framework}
\end{figure*}

Inspired by this structural alignment, we are the first to introduce this emerging paradigm to the geospatial domain, reframing the task of complex geospatial description as one of multimodally-conditioned text denoising. To operationalize this new paradigm, we present GeoDiT, a Diffusion Transformer-based vision-language model tailored for geospatial understanding. Through extensive experiments, we demonstrate that GeoDiT establishes new state-of-the-art results and validates the superiority of a structurally-aligned generative process. The principal contributions of this work are threefold:

\begin{itemize}
 \item We challenge the prevailing autoregressive paradigm by reframing geospatial language generation as a problem of resolving a parallel, spatially-grounded semantic field.
 \item We operationalize this new paradigm in GeoDiT, a diffusion transformer-based model, providing the first systematic validation that iterative parallel decoding is a structurally superior approach for generating spatially-grounded semantic descriptions in remote sensing.
 \item We provide extensive empirical evidence on a suite of challenging benchmarks, demonstrating that the performance gains achieved by GeoDiT are a direct result of its structural alignment with the data, thereby validating the superiority of the proposed generative paradigm.
\end{itemize}

%% file: sec/2_relatedwork.tex
\section{Related Work}
\subsection{Autoregressive RS Vision-Language Models}
Recent advancements in Remote-sensing Vision-Language Models (RS-VLMs) have shown remarkable potential. Pioneering works such as GeoChat~\cite{kuckreja2024geochat}, VHM~\cite{pang2025vhm}, and EarthDial~\cite{soni2025earthdial} have successfully tackled tasks from scene parsing to VQA. However, despite these architectural and data-centric innovations, they are all fundamentally built upon an autoregressive framework. This shared foundation, while powerful for general narrative tasks, imposes a rigid sequential structure on their output. Consequently, they all inherit the same structural limitation when faced with the spatially independent and non-narrative nature of geospatial intelligence—a core problem our work addresses from an entirely different paradigm.

\subsection{Diffusion-based Text Generation}
Concurrent to the advancements in autoregressive models, a distinct, non-autoregressive (NAR) generation paradigm has emerged, recently culminating in powerful diffusion-based frameworks. Rather than generating through a sequential, token-at-a-time process, these models perform generation via a parallel, iterative refinement process, typically starting from a fully masked sequence and iteratively predicting the original content over discrete timesteps. This parallel refinement approach has been successfully operationalized in end-to-end multimodal systems such as LLaDA \cite{you2025lladavlargelanguagediffusion}, LaViDA \cite{jiao2025lavidadrivevisiontextinteraction}, and MMaDA \cite{yang2025mmadamultimodallargediffusion}, which have demonstrated significant potential for unified multimodal understanding on general-domain tasks. However, these powerful NAR techniques have been predominantly developed for and evaluated on narrative-style text. Their systematic architectural adaptation for the unique structural challenges of multimodal geospatial understanding remains a critical but unexplored research gap which we aim to solve.

%% file: sec/3_methodology.tex
% \begin{figure}[!t]
%     \centering
%     \includegraphics[width=1\linewidth]{AnonymousSubmission//fig/fa.pdf}
%     \caption{xxx}
% \label{fig:framework}
% \end{figure}

\section{Methodology}
\label{sec:methodology}

Our proposed framework, GeoDiT, introduces a diffusion-based paradigm to generate structured textual descriptions of remote sensing imagery. The complete architecture, illustrated in Figure~\ref{fig:framework}, comprises a visual conditioning backbone and a core generative module, the Diffusion Transformer. In essence, the visual backbone first processes an input image to produce a set of conditioning vectors that encapsulate the image's geospatial context. These vectors are then utilized by the Diffusion Transformer, which performs a non-autoregressive, iterative decoding process to synthesize the final output. We detail the architecture of these components first, followed by a description of the training and inference procedures.

\subsection{Model Architecture}
\label{sec:architecture}

% This follows the \subsection{Model Architecture} from the previous response.
% No new subsection command is needed here.

The architecture of GeoDiT consists of two main modules: a visual backbone that provides geospatial context, and a generative core that synthesizes the textual output.

\textbf{Visual Backbone.} GeoDiT uses a pre-trained SigLIP-2 (ViT-SO400M) model~\cite{tschannen2025siglip2multilingualvisionlanguage} as its visual backbone to convert an input image $I \in \mathbb{R}^{H \times W \times 3}$ into a sequence of patch embeddings into a sequence of $N$ visual patch embeddings:

% The visual backbone of GeoDiT is responsible for converting a raw remote sensing image into a sequence of feature-rich condition vectors. This process begins with a powerful pre-trained vision encoder, specifically the SigLIP-2 (ViT-SO400M) model~\cite{tschannen2025siglip2multilingualvisionlanguage}. For an input image $I \in \mathbb{R}^{H \times W \times 3}$, the encoder transforms it into a sequence of $N$ visual patch embeddings:
\begin{equation}
    Z_v = \text{Encoder}_{\text{ViT}}(I), \quad \text{where } Z_v \in \mathbb{R}^{N \times D_v}.
\end{equation}
To align these visual features with the hidden dimension of our generative model, a lightweight Multi-Layer Perceptron (MLP) then projects $Z_v$ to produce the final visual condition vectors:
\begin{equation}
    C_v = \text{MLP}(Z_v), \quad \text{where } C_v \in \mathbb{R}^{N \times d}.
\end{equation}
Here, $D_v$ is the hidden dimension of the vision encoder, and $d$ is the hidden dimension of our core Diffusion Transformer. The sequence $C_v$ serves as the primary geospatial context for the generative process.

\textbf{Generative Core: A Modality-Adapted DiT.}
At the heart of GeoDiT lies a generative module designed to synthesize structured text. We adopt the core architectural principle of Diffusion Transformers (DiT)~\cite{hatamizadeh2024diffit}: leveraging a standard Transformer as a powerful, scalable backbone for a diffusion-based generative process. However, a fundamental adaptation is required to accommodate our target modality. While the original DiT operates on continuous latent variables via Gaussian diffusion, our objective is to generate discrete text tokens. We therefore reformulate the generative core to operate within a discrete, mask-and-predict diffusion framework.

Our generative core is a bidirectional Transformer, denoted as $p_\theta$, which learns to reverse a forward masking process. Architecturally, its primary function is to predict the original identities of masked tokens in a sequence, conditioned on both unmasked textual context and the visual condition vectors $C_v$. At each reverse diffusion step $t$, the input to the Transformer is a hybrid sequence $X_t$ formed by concatenating the visual vectors with the embeddings of the partially masked text sequence $T_t$:
\begin{equation}
    X_t = \text{concat}\left( C_v, \mathbf{E}(T_t) \right)
    \label{eq:input_concat},
\end{equation}
where $\mathbf{E}$ is the text token embedding layer. This sequence is then processed by the stack of $L$ Transformer blocks:
\begin{equation}
    H_t = \text{Transformer}_\theta(X_t)
    \label{eq:transformer_core}.
\end{equation}
% The resulting hidden states corresponding to the text tokens, $H_t^{\text{text}}$, are then projected to produce a probability distribution over the entire vocabulary $V$ for each masked position:
From the full sequence of output hidden states $H_t$, we select only those corresponding to the text tokens, denoted as  $H_t^{\text{text}}$. These are then projected to produce a probability distribution over the entire vocabulary $V$ for each masked position:
\begin{equation}
    p_\theta(T_0 | T_t, C_v) = \text{softmax}(\mathbf{W}_p H_t^{\text{text}} + \mathbf{b}_p)
    \label{eq:output_projection},
\end{equation}
where $\mathbf{W}_p$ and $\mathbf{b}_p$ are the parameters of the final linear projection layer.

% This formulation directly adapts the DiT blueprint. The Transformer backbone, $\text{Transformer}_\theta$, remains the central, scalable component responsible for modeling complex dependencies. However, its predictive objective is systematically shifted from regressing a continuous noise vector (as in the original DiT) to classifying discrete tokens, a necessary adaptation for generative modeling in the text modality. This non-autoregressive, bidirectional architecture allows the model to utilize the full visual and partial textual context simultaneously to make globally coherent predictions. Emperically, to endow our generative core with strong foundational language capabilities, we initialize its weights from the pre-trained LLaDA-8B model~\cite{nie2025largelanguagediffusionmodels}. LLaDA, being a powerful diffusion-based language model with a compatible bidirectional Transformer architecture, serves as an ideal starting point for our domain-specific adaptation.

In operationalizing this architecture, we identified the recent LLaDA-8B model~\cite{nie2025largelanguagediffusionmodels} as a particularly suitable foundation. While not explicitly described as a Diffusion Transformer in its original presentation, its core mechanism, a bidirectional Transformer optimized for iterative mask-and-predict tasks,  is fundamentally aligned with the discrete diffusion principles we require. This fundamental alignment establishes LLaDA-8B as a direct, pre-trained implementation of our proposed paradigm, whose bidirectional and iterative nature we have argued is essential for resolving the parallel semantics of geospatial data. Therefore, we adopt this validated architecture for GeoDiT, initializing our generative core with the publicly available LLaDA-8B weights to leverage its powerful foundational capabilities. Consequently, the primary novelty of our work lies not in redesigning this foundational architecture, but in the systematic methodology we developed to ground its powerful generative capabilities in the unique, non-narrative semantics of the geospatial domain.

\begin{table}[!t]
\renewcommand{\arraystretch}{1.6}
% \vspace{-0.3cm}
\resizebox{0.98\linewidth}{!}{
\fontsize{10pt}{10pt}\selectfont
\begin{tabular}{cccc}
\hline
Task                                    & Data                & Size & Type   \\ \hline
\multirow{5}{*}{\begin{tabular}[c]{@{}c@{}}Image \\ Captioning\end{tabular}}          & RSICD  \cite{lu2017exploring}                    & 24,333& optical  \\ \cline{2-4} 
                                        & UCM-Captions \cite{cheng2022nwpu}               & 9,986& optical  \\ \cline{2-4} 
                                        & RSITMD  \cite{yuan2022exploring}                   & 20,096& optical  \\ \cline{2-4} 
                                        & Sydney-Captions \cite{qu2016deep}             & 2,837& optical  \\ \cline{2-4} 
                                        & NWPU-Captions \cite{cheng2017remote}             & 141,631& optical  \\ \hline
\multirow{6}{*}{\begin{tabular}[c]{@{}c@{}}VQA \\ \\ \vspace{-1.5ex} \end{tabular}} & FloodNet \cite{rahnemoonfar2021floodnet}                    & 4,511& optical  \\ \cline{2-4} 
                                        & RSVQA$\_$LR \cite{lobry2020rsvqa}              & 67,228& optical  \\ \cline{2-4} 
                                        & RSIVQA \cite{zheng2021mutual}                    & 68,625& optical  \\ \cline{2-4} 
                                        & CRSVQA \cite{zhang2023multistep}                    & 900& optical  \\ \hline
\multirow{7}{*}{Classification}         & NWPU-RESISC45 \cite{cheng2022nwpu}              & 6,300& optical  \\ \cline{2-4} 
                                        & EuroSAT \cite{helber2019eurosat}                    & 5,400& optical  \\ \cline{2-4} 
                                        & UCMerced-Landuse \cite{yang2010bag}       & 420& optical  \\ \cline{2-4} 
                                        & WHU-RS19 \cite{dai2010satellite}                 & 196& optical  \\ \cline{2-4} 
                                        & RSSCN7 \cite{zou2015deep}                    & 560& optical  \\ \cline{2-4} 
                                        & DSCR \cite{di2021public}                      & 11,951& optical  \\ \cline{2-4} 
                                        & FGSCR-42 \cite{di2019public}                     & 3,878& optical  \\ \hline
\multirow{8}{*}{Detection}             & DOTA \cite{ding2021object}                  & 163,486& optical  \\ \cline{2-4} 
                                        & DIOR  \cite{han2014object}                      & 58,200& optical  \\ \cline{2-4} 
                                        & FAIR1M \cite{sun2022fair1m}                      & 40,466& optical  \\ \cline{2-4} 
                                        & NWPUVHR10 \cite{li2017rotation}                  & 3,190& optical  \\ \cline{2-4} 
                                        & HRRSD \cite{zhang2019hierarchical}                     & 23,044& optical  \\ \cline{2-4} 
                                        & RSOD \cite{long2017accurate}                      & 747& optical  \\ \cline{2-4} 
                                        & UCAS-AOD \cite{zhu2015orientation}                      & 1,510& optical  \\ \cline{2-4}  
                                         \hline
Visual Grounding                        & DIOR-RSVG \cite{zhan2023rsvg}                 & 30,820& optical  \\ \hline
Region-level Captioning                        & DIOR-RSVG \cite{zhan2023rsvg}                  & 30,820& optical  \\ \hline
\end{tabular}
}
\caption{Details on the datasets collected from the MMRS-1M.}
\label{tab:constituent_datasets}
\vspace{-0.5cm}
\end{table}

\subsection{Training Strategy}
\label{sec:training}

The training of GeoDiT is guided by a unified training objective, applied across a systematic, two-stage process. This approach is designed to first establish a fundamental vision-language alignment and then cultivate advanced instruction-following capabilities for the remote sensing domain.

\begin{figure}[!t]
    \centering
    \includegraphics[width=1\linewidth]{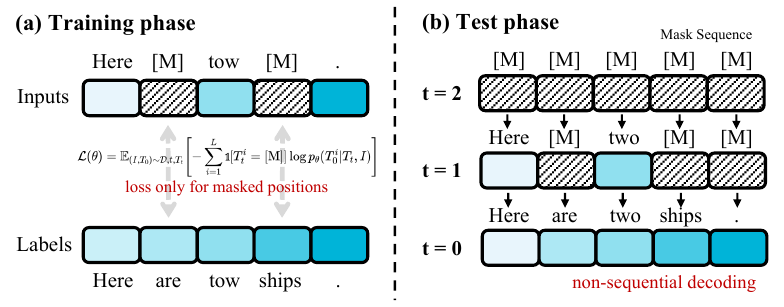}
    \caption{Illustration of the core mask-and-predict mechanism. (a) Training phase: The model is optimized to reconstruct the original text Labels from a randomly masked version Inputs. The loss is calculated only on the tokens that were masked. (b) Test phase: Generation is a non-sequential, iterative process. It starts from a fully masked sequence ($t=2$) and progressively predicts the full sentence over a series of refinement steps ($t=1$, $t=0$).}
    \vspace{-0.5cm}
\label{fig:two-phases}
\end{figure}

\textbf{Training Objective.}
The core objective is to train the parameters $\theta$ of the generative Transformer, $p_\theta$, to reverse a discrete diffusion process. This process is formulated as a mask-and-predict task. Given a ground-truth text description $T_0 = (T_0^1, \dots, T_0^L)$ and its corresponding image $I$, the training proceeds as follows. First, a random timestep $t \sim U[0, 1]$ is sampled. A forward process $q$ then generates a corrupted version $T_t$ by replacing each token $T_0^i$ with a special `[M]' token, independently and with probability $t$.

The model $p_\theta$ is then optimized to predict the original tokens $T_0$ given the corrupted sequence $T_t$ and the visual context from image $I$. This is achieved by minimizing the following loss function, $\mathcal{L}(\theta)$, which is an upper bound on the negative log-likelihood of the data:
\begin{equation}
    \begin{aligned}
        &\mathcal{L}(\theta) = \\
        & \mathbb{E}_{(I, T_0) \sim \mathcal{D}} \left[ -\sum_{i=1}^{L} \mathds{1}[T_t^i = \text{[M]}] \log p_\theta(T_0^i | T_t, I) \right],
    \end{aligned}
    \label{eq:loss_function}
\end{equation}
where $\mathcal{D}$ is the training dataset, $\mathds{1}[\cdot]$ is the indicator function which activates the loss only for masked positions, and $p_\theta(T_0^i | T_t, I)$ is the probability assigned by the model to the correct token $T_0^i$ at position $i$. This objective effectively trains the model to denoise the text by filling in the masked tokens based on multimodal context, as visually summarized in Figure~\ref{fig:two-phases}(a).

\begin{table*}[!tb]
\begin{center}
\vspace{-0.3cm}
\resizebox{\textwidth}{!}{
\begin{tabular}{lcccccccccccc} % 确保这里有19列
\toprule
\multirow{2}{*}{Method} &\multicolumn{3}{c}{\textbf{UCM-Captions}} & \multicolumn{3}{c}{\textbf{RSICD}} & \multicolumn{3}{c}{\textbf{NWPU-Captions}} & \multicolumn{3}{c}{\textbf{Sydney-Captions}}\\
\cmidrule(lr){2-4} \cmidrule(lr){5-7} \cmidrule(lr){8-10} \cmidrule(lr){11-13}
 &\textbf{B-4} \faCommentDots[regular] & \textbf{MT} \faCommentDots[regular] & \textbf{Cr} \faCrosshairs
 &\textbf{B-4} \faCommentDots[regular] & \textbf{MT} \faCommentDots[regular] & \textbf{Cr} \faCrosshairs
 &\textbf{B-4} \faCommentDots[regular] & \textbf{MT} \faCommentDots[regular] & \textbf{Cr} \faCrosshairs
 &\textbf{B-4} \faCommentDots[regular] & \textbf{MT} \faCommentDots[regular] & \textbf{Cr} \faCrosshairs \\
\midrule
\multicolumn{13}{l}{\textcolor{gray}{\textit{Commercial Autoregressive Vision-Language Models}}} \\
GPT-4V~\cite{achiam2023gpt} &7.4&18.7& 34.8&5.4& 14.8& 30.9& 15.2& 22.9& 45.6& 9.1& 18.5& 24.0 \\
Claude-4 (Anthropic 2025)  &4.3 & 11.7 & 22.2 &2.1 &10.6 &19.8 &11.6 &18.9 &42.8 &5.7& 15.5 &21.6\\
\midrule
\multicolumn{13}{l}{\textcolor{gray}{\textit{Open-source Diffusion-based Vision-Language Models}}} \\
LLaDA-V ~\cite{you2025lladavlargelanguagediffusion}  & 22.8 & 17.8 & 38.6 &9.8 &10.9 &31.4 &12.3 &13.6 &29.8 &16.4 &12.8 & 24.3\\
LaVida~\cite{li2025lavidalargediffusionlanguage}  & 24.0& 16.2& 42.4& 12.5& 11.2&30.1& 15.0& 16.9& 33.8&19.4& 16.7& 32.9\\
MMaDA~\cite{yang2025mmadamultimodallargediffusion}  & 25.6 & 18.9 & 43.6 & 15.8 & 13.9 & 31.2 & 14.6 & 15.7 & 31.1 & 20.2 & 19.6 & 32.8\\
\midrule
\multicolumn{13}{l}{\textcolor{gray}{\textit{Open-source Autoregressive Vision-Language Models}}} \\
LLaVA-1.5~\cite{liu2024improvedbaselinesvisualinstruction}  & 24.6 & 26.7 &47.7 &21.3 &16.4 &43.6 &47.6 &\underline{26.5} &56.9 &39.1 &25.3 &22.6 \\
Qwen2.5-VL~\cite{bai2025qwen25vltechnicalreport} & 23.1 & \underline{30.5} &35.9 &13.9 &20.6 &35.3 &24.1 &26.4 &47.8 &16.1 & 22.5 &27.9  \\
GeoChat~\cite{kuckreja2024geochat} &21.0 &20.8 &36.8 &15.8 &14.1 &34.4 &37.8 &25.2 &44.2 &24.6 &16.5 &20.4 \\
VHM~\cite{pang2025vhm}& \underline{42.4} &26.9 &\underline{64.2} &19.5 &22.2 &53.5 &\underline{47.9} &18.9 &40.4 &37.0 &21.5 &33.2 \\
EarthDial~\cite{soni2025earthdial} &14.6 &18.1 &35.9 &\underline{27.8} &\underline{25.4} &\underline{115.3} &35.9 &28.8 &\underline{69.3} &\underline{45.0} &\underline{39.4} &\underline{113.0}\\
\midrule
 \textbf{GeoDiT (ours)} &\textbf{44.7}& \textbf{32.9}&\textbf{73.8} &\textbf{28.6}&\textbf{26.8}&\textbf{135.6} &\textbf{62.2}&\textbf{28.9}&\textbf{77.4} &\textbf{47.2}&\textbf{40.8}&\textbf{128.3}\\
\bottomrule
\end{tabular}
}
\end{center}
\vspace{-0.3cm}
\caption{Quantitative comparison on remote sensing captioning benchmarks. Metrics are grouped to assess distinct qualities: narrative-based  (BLEU4 \faCommentDots[regular], METEOR \faCommentDots[regular]) and object-centric (CIDEr \faCrosshairs). Best performance is in bold. Note GeoDiT's significant and consistent advantage on the object-centric CIDEr metric.}
\label{tab:ic}
\vspace{-0.5cm}
\end{table*}

\textbf{Stage \uppercase\expandafter{\romannumeral1}: Vision-Language Alignment.} We freeze the vision encoder and generative Transformer, training only the lightweight MLP projector to align the visual and text embedding spaces by optimizing the objective function in Eq.~\eqref{eq:loss_function}.  We utilize SkyScript dataset~\cite{wang2024skyscript} in this stage. This step effectively teaches the projector to map visual concepts into a representation that the generative core can readily understand, without the computational expense of updating the large backbone models.

\textbf{Stage \uppercase\expandafter{\romannumeral2}: Full Instruction Tuning.}
We unfreeze all model components and perform end-to-end training on the optical subset of the MMRS-1M dataset shown in Table~\ref{tab:constituent_datasets}, created by amalgamating 34 remote sensing datasets into a unified instruction-following format. By minimizing the loss $\mathcal{L}(\theta)$ from Eq.~\eqref{eq:loss_function}, this end-to-end tuning allows all components to co-adapt, specializing the model for the nuanced task of remote sensing interpretation.

\subsection{Inference Procedure}
\label{sec:inference}

GeoDiT employs a non-autoregressive, iterative process that synthesizes the entire textual description by progressively denoising a masked sequence into the final response. This approach, illustrated in Figure~\ref{fig:two-phases}(b), directly mirrors the reverse of the diffusion process learned during training.

Generation begins with a fully masked text template $T_{t=N} = \{[M], \dots, [M]\}$ of a predetermined length $L$, corresponding to the initial timestep $t_N=1$. The model then iteratively refines this template over a series of $N$ discrete timesteps $(t_N, t_{N-1}, \dots, t_1)$, where $t_1 \approx 0$. At each iteration $k$ (from timestep $t_k$ to $t_{k-1}$), the generative core $p_\theta$ first produces a complete prediction of the final text, denoted $\hat{T}_0$, by taking the most likely token at each position from the model's output distribution:
\begin{equation}
    \hat{T}_0 = \underset{T'_0}{\text{argmax}} \ p_\theta(T'_0 | T_{t_k}, C_v),
    \label{eq:inference_prediction}
\end{equation}
where $T_{t_k}$ is the masked text sequence at the current step.

This complete prediction, $\hat{T}_0$, serves as an intermediate estimate. To construct the input for the next, less-masked iteration, $T_{t_{k-1}}$, a scheduled remasking is applied. A key component of this process is the remasking strategy, which determines which tokens to reveal and which to refine. Following the effective approach in discrete diffusion models~\cite{nie2025largelanguagediffusionmodels}, we employ a \textbf{low-confidence remasking} strategy. Based on the confidence scores from the output probabilities $p_\theta(\cdot | T_{t_k}, C_v)$, we preserve the tokens the model is most certain about from $\hat{T}_0$, and revert the less certain ones back to the `[M]' token to form $T_{t_{k-1}}$.

This iterative refinement of prediction and remasking continues until the final timestep is reached. The resulting sequence at this point, now fully "denoised", is produced as the final textual description. This method allows GeoDiT to consider the global context of both the image and the entire text sequence at every step, leading to more coherent and globally consistent outputs.

\begin{table}[!t]
\begin{center}
%\vspace{-0.3cm}
\resizebox{0.98\linewidth}{!}{
\begin{tabular}{lcccccccc} % 18 columns now
\toprule
\multirow{2}{*}{\textbf{Method}}&  \multicolumn{2}{c}{\textbf{DIOR-RSVG}} 
& \multicolumn{2}{c}{\textbf{VRSBench}} 
& \multicolumn{2}{c}{\textbf{AVVG}} 
& \multicolumn{2}{c}{\textbf{RSVG}} 
 \\
\cmidrule(lr){2-3} \cmidrule(lr){4-5} \cmidrule(lr){6-7} \cmidrule(lr){8-9}
& \textbf{VG} & \textbf{DET} 
& \textbf{VG} & \textbf{DET}  
& \textbf{VG} & \textbf{DET}  
& \textbf{VG} & \textbf{DET}   \\
\midrule
\multicolumn{9}{l}{\textcolor{gray}{\textit{Commercial Autoregressive Vision-Language Models}}} \\
GPT-4V & 3.7 & - & 0 & - & 0 &- & 1.5 &-   \\
Claude-4 &9.0 &- &0.1 &- &0 &- &2.0 &- \\
\midrule
\multicolumn{9}{l}{\textcolor{gray}{\textit{Open-source Diffusion-based Vision-Language Models}}} \\
LLaDA-V & 0.0 &- & 0.0 &- & 0.0 &- & 0.0 &- \\
LaVida & 0.0 &- & 0.0 &- & 0.0 &- & 0.0 &-  \\
MMaDA  & 0.0 &- & 0.0 &- & 0.0 &- & 0.0 &- \\
\midrule
\multicolumn{9}{l}{\textcolor{gray}{\textit{Open-source Autoregressive Vision-Language Models}}} \\
LLaVA-1.5 & 11.4 &2.2 & 15.4 &3.8 & 0.5 &0.0 & 10.5 & 2.8 \\
Qwen2.5-VL & 36.3 &\underline{17.9} & 45.2 &\underline{19.6} & 0.5 &0.0 & 1.0 & 0.0 \\
GeoChat & 31.4 &1.4 & \underline{56.3} &3.4 & \underline{7.5} &\underline{0.4} & 5.5 & 0.0 \\
VHM & \underline{55.9} &3.4 & 33.9 &2.8 & 0.0 &0.0 & 2.5 & 0.0\\
EarthDial & 46.1 &13.1 & 14.4 &0.7 & 3.5 &0.0 & \underline{42.0} & \underline{13.5} \\
\midrule
\textbf{GeoDiT (Ours)} & \textbf{60.4} &\textbf{20.8} & \textbf{63.7} &\textbf{24.9} & \textbf{21.7} & \textbf{11.4}  &\textbf{43.2} & \textbf{18.7}  \\
\bottomrule
\end{tabular}
}
\end{center}
\vspace{-0.3cm}
\caption{ Quantitative comparison on visual grounding (VG) and object detection (DET) benchmarks.}
\label{tab:vg}
\vspace{-0.6cm}
\end{table}

\subsection{Implementation Details}
\label{sec:implementation_details}

Our framework is implemented using the PyTorch library and trained on NVIDIA H200 GPUs. Below we detail the specific configurations and hyperparameters for our model's components, training, and inference stages.

\textbf{Model Configurations.}
The key components of GeoDiT are configured as follows. 
The Visual Backbone is a SigLIP-2 (ViT-SO400M) model, which outputs visual embeddings with a hidden dimension of $D_v = 1152$.
The MLP Projector consists of two linear layers with a GELU activation function in between, projecting the visual embeddings from $D_v$ to the generative core's hidden dimension $d$.
Our Generative Core, initialized from LLaDA-8B, is a 32-layer bidirectional Transformer with a hidden dimension of $d=4096$ and 32 attention heads.

\textbf{Training Details.}
The two-stage training process utilizes the AdamW optimizer with $\beta_1=0.9$ and $\beta_2=0.95$.
For Stage \uppercase\expandafter{\romannumeral1} (Vision-Language Alignment), we train the MLP projector for 1 epoch on the SkyScript \cite{wang2024skyscript} dataset. The global batch size is 96. We employ a cosine learning rate scheduler, which includes a warm-up phase over the first 3\% of the total training steps to a peak learning rate of $1 \times 10^{-3}$, after which the learning rate decays. No weight decay is applied in this stage.
For Stage \uppercase\expandafter{\romannumeral2} (Full Instruction Tuning), the entire model is fine-tuned for 1 epochs on our curated MMRS-1M \cite{zhang2024earthgpt} optical subset. The global batch size for this stage is 24. We again utilize a cosine learning rate scheduler with a warm-up phase over the first 3\% of training steps, but with a lower peak learning rate of $1 \times 10^{-5}$. No weight decay is used.

\textbf{Inference Details.}
During inference, the iterative generation process is performed over $N=8$ discrete timesteps. The maximum length $L$ for the generated response is set to 16 tokens for image captioning, 32 tokens for object detection and 8 tokens for others. We employ the low-confidence remasking strategy for iterative refinement. All reported results are generated using greedy decoding and without classifier-free guidance.

%% file: sec/4_experiments.tex
\section{Experiments}
This section presents a comprehensive set of experiments designed to validate the central hypothesis of this work: that reframing geospatial language generation as a parallel, iterative refinement process offers a structurally superior alternative to the conventional autoregressive paradigm. To this end, we first detail the experimental setup, including the datasets, evaluation metrics, and baselines (Section~\ref{sec:setup}). We then present the main quantitative results on a diverse suite of tasks, including image captioning , visual grounding , and visual question answering  (Section~\ref{sec:results}). Following this, we conduct a series of ablation studies to dissect the model and analyze the contribution of its key components (Section~\ref{sec:abla}). Finally, we provide a qualitative analysis with visual examples to offer further intuitive insights into the model's behavior and advantages (Section~\ref{sec:vis}).

\begin{table}[!tb]
\begin{center}
	\resizebox{0.98\linewidth}{!}{
	\begin{tabular}{lcccccccc}
		\toprule
        \multirow{2}{*}{\textbf{Method}} & \multicolumn{3}{c}{\textbf{RSVQA-LR}} & \multicolumn{2}{c}{\textbf{RSVQA-HR}} & \multicolumn{2}{c}{\textbf{Classification}} \\
        \cmidrule(lr){2-4} \cmidrule(lr){5-6} \cmidrule(lr){7-8}
		 & \textbf{LR-R} & \textbf{LR-P} & \textbf{LR-C} & \textbf{HR-A} & \textbf{HR-C} & \textbf{AID} & \textbf{WHU-RS19}\\
\midrule
\multicolumn{6}{l}{\textcolor{gray}{\textit{Commercial Autoregressive Vision-Language Models}}} \\
GPT-4V &74.3  &81.3 &56.7 &18.4 &33.6 & 77.3 & \underline{94.7} \\
Claude-4 &63.4 &79.6 &65.4 &21.7 &42.8 & 45.5 & 66.1 \\
\midrule
\multicolumn{6}{l}{\textcolor{gray}{\textit{Open-source Diffusion-based Vision-Language Models}}} \\
LLaDA-V & 70.3 & 62.7 &18.6 & 36.3 & 34.8 &56.4 &62.3\\
LaVida & 66.8 &58.1 &14.2 &26.6 &44.3 &55.8 &60.8\\
MMaDA & 71.6 &60.8 &16.8 & 28.9 & 37.6 &57.6 &59.7\\
\midrule
		\multicolumn{6}{l}{\textcolor{gray}{\textit{Open-source Autoregressive Vision-Language Models}}} \\
        LLaVA-1.5    & 75.5 & 57.1 & 70.6   & 0  & 66.0 &47.3 &62.1 \\
		Qwen2.5-VL   & 94.3 & 62.5 & 43.1   & 0  & 53.4 &63.8 &79.5 \\
		GeoChat  & \underline{96.2} & \underline{89.3} & \underline{88.2}  & 23.3 & 75.7 &71.5 &89.5 \\
		VHM  & 90.6 & 80.4 & 86.3  & 24.3 & \underline{79.6} &\underline{79.0} &91.8 \\
EarthDial & 66.0 & 64.3 & 87.3  & \underline{34.9} & 71.8 &62.5 &73.4 \\
\midrule
		\textbf{GeoDiT (ours)} &\textbf{98.1} & \textbf{91.1} & \textbf{90.2}  & \textbf{37.6} & \textbf{80.6} & \textbf{81.2} & \textbf{95.0}\\
		\bottomrule
	\end{tabular}
}
\end{center}
\vspace{-0.3cm}
% \begin{flushleft}
% \tiny \textit{R/P/C/A/D: Rural / Presence / Comparison / Area / Direction.}
% \end{flushleft}
\begin{tablenotes}[flushleft]
            % \item[] \tiny\textit{LR/HR: RSVQA-LR/RSVQA-HR}
            \item[] \tiny\textit{R/P/C/A: Rural / Presence / Comparison / Area }
        \end{tablenotes}
% \vspace{-0.2cm}
\caption{Performance evaluation on remote sensing visual question answering (RSVQA) and scene classification. }
\label{tab:vqa}
\vspace{-0.6cm}
\end{table}

\subsection{Experimental Setup}
\label{sec:setup}
\textbf{Datasets} We evaluate our model on a wide array of public benchmarks to assess its performance across diverse geospatial tasks. For remote sensing captioning, we use four standard datasets: UCM-Captions~\cite{qu2016deep}, RSICD~\cite{lu2017exploring}, NWPU-Captions~\cite{cheng2022nwpu}, and Sydney-Captions~\cite{qu2016deep}. For visual grounding and object detection, our evaluation is conducted on DIOR-RSVG~\cite{zhan2023rsvg}, VRSBench~\cite{NEURIPS2024_05b7f821}, AVVG~\cite{zhou2024geoground}, and RSVG~\cite{sun2022visual}. For visual question answering and scene classification, we employ the RSVQA-LR/HR~\cite{lobry2020rsvqa}, AID~\cite{xia2017aid}, and WHU-RS19~\cite{xia2010structural} benchmarks. This comprehensive suite of benchmarks ensures a thorough and robust evaluation of GeoDiT's capabilities and its generalization across different tasks and data domains.

\textbf{Evaluation Metrics.} We employ a set of standard metrics to evaluate performance across the different tasks. For captioning, we report scores for BLEU-4 (B-4), METEOR (MT), and CIDEr (Cr). We place particular emphasis on the CIDEr score, as it is specifically designed to measure consensus in object-centric descriptions, which directly aligns with our core hypothesis regarding the generation of structured, non-narrative semantics. With an IoU threshold of 0.5, we use Accuracy (Acc@0.5) for visual grounding (VG) task and mean Average Precision (mAP@0.5) for object detection (DET) task. For all visual question answering and classification tasks, we generate responses using greedy decoding and report accuracy ultilizing BERT-based similarity.

\textbf{Compared Methods.} We conduct a comprehensive comparison of GeoDiT against a wide range of state-of-the-art vision-language models. To provide a clear overview of the competitive landscape, we group these baselines into three main categories: 1)~\textbf{Commercial Autoregressive Models}: Large-scale, proprietary systems such as GPT-4V and Claude-4 ~\cite{enis2024llmnmtadvancinglowresource} that represent the upper echelon of general VLM capabilities. 2)~\textbf{Open-source Diffusion-based Models}: This category includes LLaDA-V, LaVida, and MMaDA. While they operate on similar non-autoregressive principles, they are designed for general-domain tasks and not specifically architected for geospatial semantics. 3)~\textbf{Open-source Autoregressive Models}: This group comprises the dominant paradigm in academic research and includes leading remote sensing VLMs like LLaVA-1.5, Qwen2.5-VL, GeoChat, VHM, and EarthDial. This curated set of baselines allows for a multi-faceted evaluation of our model against both structurally similar and fundamentally different state-of-the-art approaches.

% \paragraph{Implementation Details} The detailed configurations and hyperparameters for our model and training procedures are described in the \nameref{sec:implementation_details} subsection.

\subsection{Main Results}
\label{sec:results}
We now present the main results of our experiments, evaluating GeoDiT's performance on a diverse set of tasks and comparing it against state-of-the-art models.

\textbf{Image Captioning.} The results for remote sensing image captioning are presented in Table~\ref{tab:ic}. On the evaluated benchmarks, GeoDiT establishes a new state-of-the-art for the BLEU-4, METEOR, and CIDEr metrics. For instance, on the RSICD benchmark, GeoDiT achieves a CIDEr score of 135.6, marking a 17.6\% relative improvement over the strongest competitor (EarthDial). Similarly, on Sydney-Captions, our model's score of 128.3 represents a 13.5\% relative gain over the runner-up. This consistent and dramatic outperformance on CIDEr, a metric specifically designed to evaluate consensus on the set of contained objects, provides powerful validation for our thesis, highlighting a fundamental alignment between GeoDiT's parallel refinement process and the nature of unordered geospatial descriptions.

\begin{figure}
    \centering
    \includegraphics[width=1\linewidth]{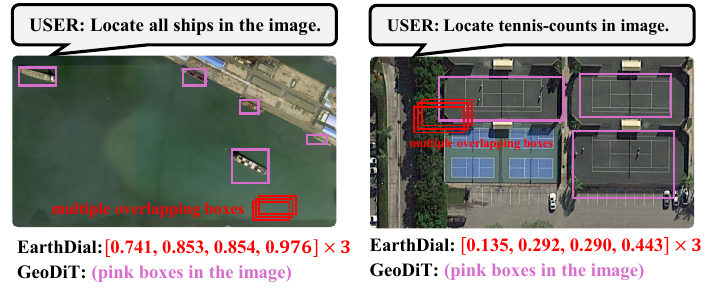}
    \caption{Qualitative examples of the generative failure mode in autoregressive models for object detection}
    \vspace{-0.5cm}
    \label{fig:det-compare}
\end{figure}

\textbf{Visual Grounding and Detection.} For tasks that demand precise spatial localization, namely visual grounding (VG) and object detection (DET), GeoDiT demonstrates a comprehensive superiority. As shown in Table~\ref{tab:vg}, GeoDiT sets a new state-of-the-art across all benchmarks and on both evaluation metrics by a significant margin. For instance, on the challenging VRSBench dataset, GeoDiT achieves a VG score of 63.7 and a DET score of 24.9, substantially outperforming the strongest baselines in each category. This gap stems from a fundamental limitation inherent to sequential generation. Autoregressive models, constrained by their token-by-token process, exhibit a critical failure mode when tasked with enumerating multiple objects. The generation process of each subsequent bounding box is conditioned on the preceding ones, creating a path dependency inducing the  premature convergence on a single salient object, manifested as the repeated generation of duplicate or minimally perturbed coordinates, precluding the identification of other distinct entities (see Fig.~\ref{fig:det-compare}). Consequently, their ability to perform a comprehensive object canvass is structurally impaired, leading to a severe degradation in DET performance.

% \textbf{Visual Question Answering and Classification.} The results for visual question answering and scene classification, presented in Table~\ref{tab:vqa}, reveal a nuanced picture of GeoDiT's capabilities. The model's primary strength lies in complex reasoning over low-resolution imagery, where it achieves new state-of-the-art performance across all three categories of the RSVQA-LR benchmark. For instance, on the RSVQA-LR-R (Rural) subtask, GeoDiT scores 98.1\%, decisively outperforming all other methods. On the more challenging high-resolution RSVQA-HR tasks, the model remains highly competitive, securing the top score of 80.6\% on the comparison subtask (HR-C) and performing on par with the leading models on the area subtask (HR-A). While its performance on standard  scene classification (AID, WHU-RS19) is robust, it does not lead the field. This pattern is informative: GeoDiT's dominance in complex VQA tasks that require understanding broad spatial relationships, coupled with its non-SOTA performance on simple classification, supports our core thesis. The model excels precisely on tasks that benefit from its structurally-aligned parallel refinement paradigm for understanding complex scenes and extracting complex relationships, rather than tasks that map an image to a single label.

\textbf{Visual Question Answering and Classification.} As presented in Table 4, GeoDiT's capabilities are demonstrated across both visual question answering and scene classification, where it establishes a new state-of-the-art on all evaluated sub-tasks and datasets.
The model's proficiency is consistent across the full suite of VQA tasks, which require parsing complex spatial relationships and object attributes. For instance, GeoDiT scores 98.1 on the RSVQA-LR-R (Rural) sub-task and 80.6 on the demanding RSVQA-HR-C (Comparison) task. This strong performance extends to scene classification, where GeoDiT also surpasses prior methods, reaching an accuracy of 95.0 on WHU-RS19.
This consistent success across a spectrum of tasks—from answering compositional queries about spatial arrangements (VQA) to holistic scene recognition (Classification)—offers a key insight into the model's underlying mechanism. This suggests that the parallel refinement paradigm provides a more fundamental advantage than being merely a specialized tool for structured outputs. By processing the scene globally and iteratively, the model excels not only at disentangling complex, localized relationships but also at forming the cohesive, scene-level understanding required for single-label classification.

\begin{figure}
    \centering
    \includegraphics[width=1\linewidth]{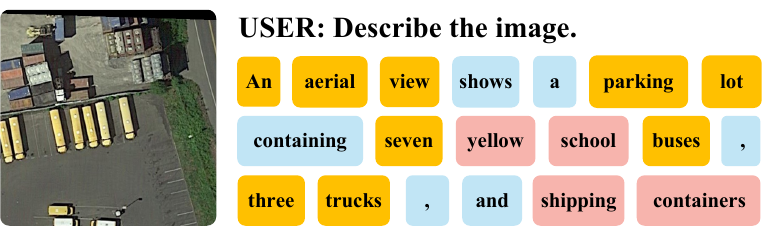}
    \caption{Visualization of the hierarchical generation process of GeoDiT. The color of each token corresponds to its relative finalization step during the iterative inference process: yellow indicates early-stage tokens, pink indicates middle-stage tokens, and blue indicates late-stage tokens. }
    \vspace{-0.7cm}
    \label{fig:process}
\end{figure}

\subsection{Qualitative Analysis}
\label{sec:vis}

To provide an intuitive understanding of the quantitative results presented in Section 4.2, we now visualize and analyze several qualitative examples. These cases are selected to highlight the practical differences in behavior between our parallel refinement paradigm and conventional autoregressive models on key geospatial tasks.

\textbf{Analysis of Structural Robustness.} Figure~\ref{fig:det-compare} provides a compelling visual demonstration of a critical failure mode inherent in autoregressive models during detection tasks. In both the harbor and tennis court scenes, the baseline model generates an initial, misplaced bounding box—for instance, locating a section of the pier instead of a ship. Critically, the model then becomes trapped in a generative loop, fixating on its initial error and repeatedly outputting the same incorrect coordinates. This behavior, visualized by the overlapping red boxes on an irrelevant location, leads to a catastrophic failure on mAP. This qualitative evidence strongly supports our central claim: the parallel refinement paradigm of GeoDiT is structurally more robust for multi-object understanding, as it avoids the sequential dependencies that lead to such error-compounding generative failures.

\textbf{Hierarchical Generation Process.} Figure~\ref{fig:process} visualizes the non-autoregressive generation process of GeoDiT. It reveals a distinct, hierarchical generation pattern that fundamentally differs from linear autoregressive models. The model first establishes the macroscopic scene context (e.g., `An aerial view shows a parking lot') and identifies the primary semantic anchors such as key objects and their counts (`seven', `buses', `three', `trucks') in the early stages (yellow). Subsequently, it refines the description by adding attributes (`yellow', `school', `shipping containers') in the middle stages (pink). Finally, it completes the sentence by inserting the necessary grammatical and syntactical components (`containing', `and', `.') in the late stages (blue). This `context-first, entity-second, syntax-last' process is only possible through a parallel, holistic understanding of the image and strongly validates our core paradigm.

\begin{table}[t!]
\centering
\resizebox{\linewidth}{!}{
\begin{tabular}{lcccc}
\toprule
\textbf{Masking Strategy} & \textbf{BLEU-4} & \textbf{CIDEr} & \textbf{mAP@0.5} & \textbf{Acc.} \\
 & (RSICD) & (RSICD) & (DIOR-RSVG) & (AID) \\
\midrule
Random Remasking      & 27.3          & 121.8          & 15.5          & 63.4          \\
Low-Confidence (Ours) & \textbf{28.6} & \textbf{135.6} & \textbf{20.8} & \textbf{67.6} \\
\midrule
Improvement (\%)      & $\uparrow$ 4.76\%       & $\uparrow$ 11.3\%       & $\uparrow$ 34.2\%       & $\uparrow$ 6.21\%       \\
\bottomrule
\end{tabular}
}
\caption{Ablation study on the masking strategy. We compare our proposed low-confidence remasking against a random remasking baseline.}
\label{tab:abla}
\vspace{-0.3cm}
\end{table}

\begin{table}[t!]
\centering
\resizebox{\linewidth}{!}{
\begin{tabular}{lcccc}
\toprule
\textbf{\#timesteps ($N$)} & \textbf{BLEU-4} & \textbf{CIDEr} & \textbf{mAP@0.5} & \textbf{Acc.} \\
 & (RSICD) & (RSICD) & (DIOR-RSVG) & (AID) \\
\midrule
1  & 21.0 & 65.8 & 7.5 & 76.5 \\
2  & 25.3 & 105.1 & 14.2 & 79.8\\
4  & 27.8 & 127.3 & 18.9 & 70.7 \\
8 & 28.6 & 135.6 & 20.8 & 81.2 \\
16 & 28.7 & 136.2 & 21.1 & 81.3 \\
\bottomrule
\end{tabular}
}
\caption{Ablation on inference timesteps ($N$). Performance saturates at $N=128$, offering the optimal trade-off between generation quality and computational cost for structured outputs.}
\label{tab:abla_n}
\vspace{-0.6cm}
\end{table}

\subsection{Ablation Studies}
\label{sec:abla}
% To validate our key design choices and understand the contribution of individual components, we conduct a series of ablation studies. We investigate three primary aspects: the impact of the iterative refinement process, the efficacy of our two-stage training strategy, and the contribution of the proposed low-confidence masking scheme.

% \paragraph{Contribution of Masking Strategy}
\textbf{Analysis of Masking Strategy.} Table~\ref{tab:abla} presents the ablation study on our masking strategy, confirming that our low-confidence approach is decisively superior to a random remasking baseline. This superiority, however, is not uniform and is most pronounced on tasks requiring high precision. We observe a remarkable 34.2\% improvement in object detection (mAP) and a significant 11.3\% gain on the object-centric CIDEr metric. In contrast, the gains are more moderate for BLEU-4 ($\uparrow$ 4.76\%) and simple VQA accuracy ( $\uparrow$ 6.21\%). This disparity is informative and expected. By focusing on uncertain tokens, our intelligent masking strategy provides the greatest benefit when refining high-stakes details like precise bounding box coordinates and key object nouns—the very elements that are critical for high mAP and CIDEr scores. This confirms our design choice is a key contributor to generating high-fidelity structured outputs.

\textbf{Analysis of Inference Timesteps.} To investigate the contribution of our iterative refinement process, we ablate the number of inference timesteps, $N$, with results detailed in Table~\ref{tab:abla_n}. A highly informative pattern emerges: performance on metrics sensitive to structural and object-centric correctness, namely CIDEr and mAP, scales steeply with the initial steps ($N=1,\cdots,16$), underscoring the necessity of a multi-step approach to resolve the parallel semantics of geospatial scenes where single-pass generation inherently falters. This trend is sharply contrasted by the far less sensitive scene classification task (Acc.), whose performance plateaus much earlier, confirming that the primary value of our method lies in meticulously constructing structured outputs rather than making a single holistic judgment. As performance across all tasks begins to saturate at $N=8$, yielding only marginal gains when doubling the computational budget to $N=16$, we adopt $N=8$ as our default setting, a robust and efficient operating point that captures the vast majority of the model's capabilities without incurring unnecessary inference latency.

%% file: sec/5_conclusion.tex
\section{Conclusion}
This work addresses a fundamental limitation of the prevailing autoregressive paradigm: its structural misalignment with the parallel nature of remote sensing data. To overcome this, we introduced GeoDiT, a novel diffusion-based vision-language model that operationalizes a parallel, iterative refinement process. Extensive experiments show GeoDiT establishes a new state-of-the-art on object-centric and structured-output tasks, such as multi-object detection and image captioning, validating the superiority of our structurally-aligned approach. Our work underscores the critical importance of aligning the generative architecture with the data's inherent structure and presents a promising new direction for complex geospatial intelligence.

%% file: sec/6_acknowledgments.tex
\section{Acknowledgments}
This work was supported by the National Natural Science Foundation of China (U22A2098, 62172185), Young Scientist Fund (32501793), China Postdoctoral Science Foundation (2025M780312),  Major Science and Technology Project of Jilin Province (20240212003GX), Major Science and Technology Project of Changchun City (2024WX05).